\documentclass[letterpaper]{article} %
\usepackage{aaai25}  %
\usepackage{times}  %
\usepackage{helvet}  %
\usepackage{courier}  %
\usepackage[hyphens]{url}  %
\usepackage{graphicx} %
\usepackage{natbib}  %
\usepackage{caption} %
\usepackage{algorithm}
\usepackage{algorithmic}

\usepackage{newfloat}
\usepackage{listings}
\DeclareCaptionStyle{ruled}{labelfont=normalfont,labelsep=colon,strut=off} %
\floatstyle{ruled}
\newfloat{listing}{tb}{lst}{}
\floatname{listing}{Listing}
\usepackage[percent]{overpic}
\usepackage{booktabs}
\usepackage{multirow}
\usepackage{amsmath}
\usepackage{amssymb}
\title{AIM: \underline{A}dditional \underline{Im}age Guided Generation of Transferable Adversarial Attacks}
\author{
    Teng Li,
    Xingjun Ma,
    Yu-Gang Jiang
}
\affiliations{
    Shanghai Key Lab of Intell. Info. Processing, School of CS, Fudan University
}

\begin{document}

\maketitle

\begin{abstract}
	Transferable adversarial examples highlight the vulnerability of deep neural networks (DNNs) to imperceptible perturbations across various real-world applications. While there have been notable advancements in untargeted transferable attacks, targeted transferable attacks remain a significant challenge. In this work, we focus on generative approaches for targeted transferable attacks.
	Current generative attacks focus on reducing overfitting to surrogate models and the source data domain, but they often overlook the importance of enhancing transferability through additional semantics.
	To address this issue, we introduce a novel plug-and-play module into the general generator architecture to enhance adversarial transferability. Specifically, we propose a \emph{Semantic Injection Module} (SIM) that utilizes the semantics contained in an additional guiding image to improve transferability. The guiding image provides a simple yet effective method to incorporate target semantics from the target class to create targeted and highly transferable attacks. Additionally, we propose new loss formulations that can integrate the semantic injection module more effectively for both targeted and untargeted attacks. We conduct comprehensive experiments under both targeted and untargeted attack settings to demonstrate the efficacy of our proposed approach.
	\begin{links}
		\link{Code}{https://terrytengli.com/s/Ce83N}
	\end{links}
\end{abstract}

\section{Introduction}
\label{sec:intro}

Over the past decades, deep neural networks (DNNs) have achieved significant success across various fields, including computer vision~\cite{alexnet} and natural language processing~\cite{lstm}. In computer vision, DNNs are widely applied to real-world tasks such as image classification~\cite{resnet,transformer}, object detection~\cite{yolo}. However, research~\cite{fgsm-attack} has demonstrated that DNNs are vulnerable to adversarial examples~\cite{intriguing}, which are modified inputs by small, imperceptible adversarial perturbations~\cite{fgsm-attack}.
Moreover, adversarial examples have been shown to transfer across different model architectures~\cite{bia-attack}, data domains~\cite{cda-attack} and modality\cite{gcma-attack,bsc-attack}. In other words, attacks crafted on one model or dataset can remain effective when applied to other models or datasets. This adversarial transferability poses a significant threat to the deployment of DNNs in real-world applications.

Transferable adversarial attacks can be crafted using various methods, which can be broadly categorized into two main types: iterative methods~\cite{pgd-attack} and generative methods~\cite{gap-attack}. Iterative attacks directly optimize the input space to generate adversarial examples, while generative attacks focus on pre-training a generator model to produce these examples. Iterative methods are often more time-consuming and may result in poorer adversarial transferability compared to generative methods, due to issues like gradient vanishing~\cite{ttta-attack}.
Attacks can also be classified based on their target: targeted attacks~\cite{ttaa-attack} and untargeted attacks~\cite{bia-attack}. Untargeted attacks aim to cause the model to predict any incorrect label, whereas targeted attacks seek to force the model to output a specific label. In the context of transferable attacks, targeted attacks are generally considered more challenging than untargeted ones~\cite{ttaa-attack}, primarily due to the risk of overfitting the surrogate model and the lack of information about the target class distribution.

The overfitting issue can be mitigated using data augmentation strategies~\cite{ucg-attack},  feature loss objectives (e.g., feature disruption~\cite{bia-attack}, batch neighborhood similarity~\cite{ttp-attack}), and unsupervised training techniques (e.g., contrastive learning~\cite{cdta-attack}). However, these methods are suboptimal because they primarily address overfitting rather than explicitly improving transferability.
Moreover, when an adversarial noise generator is trained on a specific target dataset or surrogate model architecture, the perturbations it produces may overfit to that particular context. To address this limitation, we propose to incorporate additional context-agnostic semantics to better guide the generation of transferable adversarial examples. Designing a new generator architecture for this purpose is challenging. To overcome this, we introduce the \emph{Semantic Injection Module} (SIM), a lightweight and plug-and-play module that integrates an additional guiding image into the adversarial generator, enhancing its ability to produce more transferable adversarial examples.

With SIM, we can flexibly use different guiding images to facilitate either targeted or untargeted transferable attacks. For targeted attacks, we incorporate semantic guidance from images associated with the target concept (label), improving the precision of targeted transferability. For untargeted attacks, we use guiding images from incorrect classes to help mitigate overfitting to the input image and surrogate model. Moreover, we introduce new loss formulations for adversarial loss to effectively integrate SIM into the training objectives of generative attacks.

In summary, our main contributions are:

\begin{itemize}
	\item We present a novel approach for achieving targeted transferable attacks by incorporating an additional image as guiding semantics. Specifically, we propose a lightweight plug-and-play \emph{Semantic Injection Module (SIM)}  that can be used with general adversarial generators.
	\item We investigate training objectives for generative attacks within both targeted and untargeted frameworks, including logit-level and feature-level approaches. Based on this analysis, we propose new training loss formulations that improve the effectiveness of SIM across different types of guiding semantics.
	\item We conduct extensive experiments to evaluate the efficacy of our proposed approach. Our results show that it achieves superior transferability for targeted attacks and performs on par with state-of-the-art methods for untargeted attacks.

\end{itemize}

\begin{figure*}[!t]
	\centering
	\includegraphics[width=1.0\textwidth]{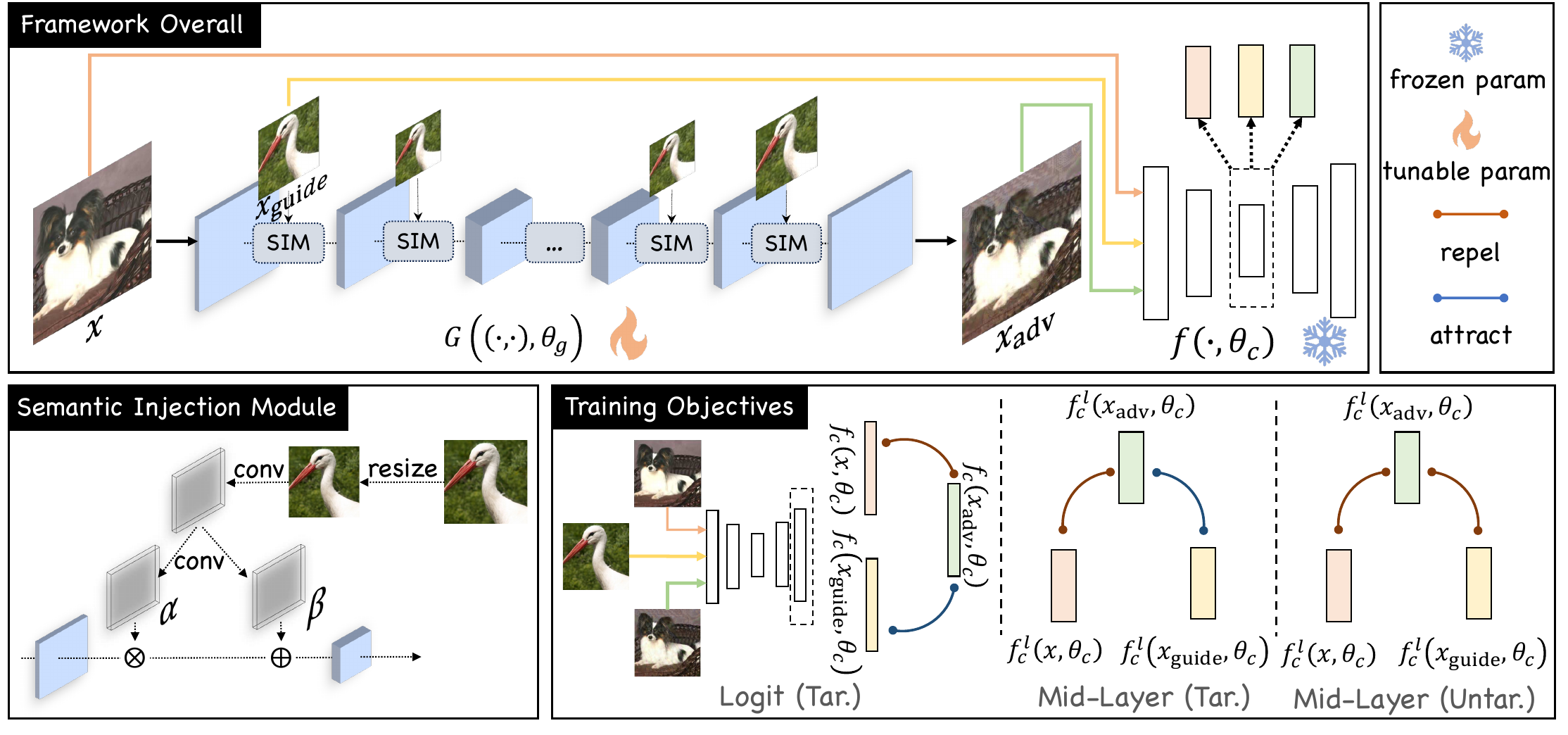}
	\caption{Our framework introduces a novel semantic injection module (SIM) into the adversarial generator $G \left( \left( \cdot, \cdot \right),\theta_{g} \right)$. The generator takes a source image $x$ and a guiding image $x_{\text{guide}}$ as inputs and outputs an adversarial example $x_{\text{adv}}$. The SIM component utilizes the feature map from the previous layer and the guiding image $x_{\text{guide}}$ to produce an enhanced feature map that incorporates the semantics from the guiding image. For targeted attacks (Tar.), we define the training objectives using logit contrastive loss and mid-layer similarity loss, which direct the adversarial example $x_{\text{adv}}$ towards the target guiding image $x_{\text{guide}}$ in both the logit and feature spaces. For untargeted attacks (Untar.), we introduce an enhanced mid-layer similarity loss to push $x_{\text{adv}}$ away from both the clean image $x$ and the guiding image $x_{\text{guide}}$ in the feature space.}
	\label{fig:framework}
\end{figure*}

\section{Related Work}
\label{sec:rela}

\subsection{Transferable Adversarial Attack}

Existing transferable adversarial attacks can be broadly classified into two types: \emph{iterative attacks} and \emph{generative attacks}.
\emph{Iterative attacks} optimize adversarial examples by constructing logits-oriented loss functions. For example, the Fast Gradient Sign Method (FGSM)~\cite{fgsm-attack} applies one-step perturbations to the input image in the direction of the input gradient. The Projected Gradient Descent (PGD) attack enhances adversarial strength through techniques such as random initialization, multi-step perturbation, and clipping~\cite{pgd-attack}. \citet{dim-attack} improved iterative FGSM's~\cite{ifgsm-attack} transferability with diverse data augmentation. Additionally, feature-space attacks, such as DR~\cite{dr-attack}, improve adversarial strength by reducing the dispersion of mid-layer features within a surrogate model. Many logits-oriented attacks can also be adapted for targeted adversarial attacks, expanding their applicability.

\emph{Generative attacks} train an adversarial generator to produce adversarial examples. In \cite{atn-attack}, a generative architecture was designed to generate adversarial examples for MNIST~\cite{lenet} images by disrupting the output logits.
In contrast to cross-model transfer attacks, CDA~\cite{cda-attack} leverages the inherent cross-domain transferability of generative models to enhance adversarial transferability across different data domains. However, establishing criteria based on logits distribution has proven inconsistent. To address this, \citet{bia-attack} expanded the adversarial objective into the feature space, disrupting the consistency of mid-layer features. Additionally, \citet{cdta-attack} approached the attack problem within a contrastive learning context, while GAMA~\cite{gama-attack} incorporated semantic supervisory signals guided by vision-language models~\cite{clip}. \citet{facl-attack} explored vulnerabilities in the image frequency domain to improve transferability. Furthermore, UCG~\cite{ucg-attack} developed a comprehensive framework by combining different techniques.
Several latest works focus on generative targeted attacks.
The TTAA framework~\cite{ttaa-attack} presents a dual discriminator architecture that enforces constraints in both the logits space and the feature space. This approach is advantageous because the feature space maintains stronger consistency across different architectures.
In addition to the global data distribution similarity matching, TTP~\cite{ttp-attack} explored batch-wise neighborhood similarity matching to integrate local neighborhood structures, thereby enhancing adversarial transferability.

\subsection{Additional Image Guided Generation}

The integration of additional image guidance into image generation and editing processes has been studied beyond the context of transferable adversarial attacks. For instance, image generators using SPADE normalization~\cite{spade} can produce highly realistic images by employing a semantic segmentation map. This framework features a novel layer that adjusts the generator's feature map based on the provided segmentation input. In image style transfer, StyleGAN~\cite{style-gan} uses the latent code of a style image to induce significant attribute shifts, which are then applied to the content latent code. Recent research in controllable image generation includes Stable Diffusion~\cite{stable-diffusion}, which utilizes a diffusion process guided by various control signals, such as textual prompts and image inputs, to generate high-quality images. In this work, we propose a novel approach to incorporating additional images for guiding the generation of targeted transferable attacks.

\section{Methodology}
\label{sec:method}

\subsection{Problem Formulation}
Given a clean image $x$ and surrogate classification model $f (\cdot,\theta_c)$ ($\theta_c$ denotes classifier parameters), the goal of a transfer adversarial attack is to craft an adversarial example $x_{\text{adv}}$ that misleads the model into predicting an incorrect label. The crafted adversarial example is then transferred to attack a target model that is of a different architecture or trained on a different dataset from the surrogate model. For untargeted attacks, the goal is to ensure that $f( x_{\text{adv}},\theta_c) \neq f( x,\theta_c)$, while adhering to the constraint $\left\| x_{\text{adv}} - x \right\|_{\infty} \leq \epsilon$, where $\epsilon$ denotes the perturbation budget. For targeted attacks, the objective shifts to $f( x_{\text{adv}},\theta_c) = y_t$, where $y_t$ is the target label specified by the adversary. Our work primarily focuses on targeted transferable attacks and takes a generative approach to improve transferability. We first train an adversarial generator $G \left( \cdot, \theta_{g} \right)$ ($\theta_g$ denotes generator parameters), based on which adversarial examples can be generated with a single forward pass of the generator: $x_{\text{adv}} = G \left( x, \theta_{g} \right)$.

\subsection{Framework Overview}
An overview of our proposed generative framework is illustrated in Figure~\ref{fig:framework}. It consists of three key components: 1) the base adversarial generator, 2) the semantic injection module, and 3) the training objectives.
We begin with the base adversarial generator, denoted as $G_{\text{base}} \left( \cdot, \theta_{g} \right)$. Similar to prior methods~\cite{bia-attack}, we utilize a ResNet generator that accepts a source image $x$ as input and produces an adversarial example $x_{\text{adv}}$.
However, different from previous approaches, we incorporate a semantic injection module into the generator to provide a lightweight plug-and-play enhancement.
This module allows us to generate $x_{\text{adv}}$ using additional semantics information from a guiding image $x_{\text{guide}}$. By integrating the semantic injection module, the enhanced generator can now accept two inputs: the source image $x$ and the guidance image $x_{\text{guide}}$. This enables us to formulate the generation process as $x_{\text{adv}} = G((x, x_{\text{guide}}), \theta_{g})$. To ensure the seamless integration of the semantic injection module into the adversarial generator, we need new and more advanced training objectives. For targeted attacks, we employ the logit contrastive loss and mid-layer similarity loss as the training objectives to ensure logit- and feature-level transferability. This is because targeted attacks generally need more precise guiding information toward the target label when transferability is concerned.
On the other hand, for untargeted attacks, we introduce an enhanced mid-layer similarity loss to entire feature transferability. This is because feature disruption is enough to cause errors in the target model.
Next, we will introduce the two components proposed in this work: \emph{semantic injection module} and \emph{training objectives}.

\subsection{Semantic Injection Module}

Previous transfer attacks have primarily concentrated on improving training mechanisms~\cite{cdta-attack}, designing higher-dimensional loss objectives~\cite{bia-attack}, and mining frequency-based data properties~\cite{facl-attack} to reduce the risk of overfitting in generation.
However, the transferability of these methods is inherently constrained by the capability of the generator, i.e., the generator does not have the ability to know what or where to transfer.
In other words, training an adversarial generator on a specific dataset or surrogate model can lead to context-specific overfitting.
Intuitively, leveraging additional semantics about the target class (or incorrect classes) as external guidance can enhance transferability across models or domains.
However, integrating this additional guidance into an adversarial generator poses a significant challenge.

To tackle the above challenge, we introduce a lightweight semantic injection module, which is designed to seamlessly integrate the semantics guidance provided by an additional image into the adversarial generator.
This module serves as a plug-and-play component that can be easily incorporated into the commonly used base generators. As depicted at the bottom left of Figure~\ref{fig:framework}, the semantic injection module specifically focuses on extracting and injecting semantic information into the intermediate layers of the adversarial generator.
It utilizes an affine transformation on the generator's feature map to modify the semantic attributes. The affine transformation is defined by two learnable parameters: 1) the scale parameter $\alpha$, which adjusts the feature map; and 2) the shift parameter $\beta$, which translates the feature map. Formally, the transformation is defined as:
\begin{equation}
	\label{eq:si}
	\small
	\begin{cases}
		 & f_{\texttt{SIM}}^{i} = \left( 1 + \alpha_{i} \right) f^{i} + \beta_{i},    \\
		 & \alpha_{i} = \texttt{Conv}\left( x_{g}^{i} \right),                        \\
		 & \beta_{i}  = \texttt{Conv}\left( x_{g}^{i} \right),                        \\
		 & x_{g}^{i}  = \texttt{Interp}\left( x_{\text{guide}}, w_{i}, h_{i} \right), \\
		 & i          = 1,2,\cdots,N_{\texttt{SIM}},
	\end{cases}
\end{equation}
where $f^{i}$ and $f_{\texttt{SIM}}^{i}$ represent the input and output feature maps of the $i$-th semantic injection module, respectively; the scale parameter $\alpha_{i}$ and the shift parameter $\beta_{i}$ are learnable parameters with semantic guidance; $x_{g}$ denotes the resized guided image; $\texttt{Conv}$ and $\texttt{Interp}$ denote the convolutional operation and the interpolation operation, respectively; $N_{\texttt{SIM}}$ denotes the total number of semantic injection modules.

The guiding image can be flexibly selected according to the attack goal, i.e., targeted or untargeted.
In the case of targeted attacks, we use a randomly selected image of the target concept (class) as the guiding image $x_{\text{guide}}$. As the adversary knows its target label, such image can be easily collected from the same source data domain as the input image $x$ or using image search engines like Google Images.
For untargeted attacks, $x_{\text{guide}}$ can be randomly selected from an arbitrary incorrect class. As the adversary also knows the correct class of clean image $x$, this selection can also be easily done following the same strategy as the targeted case. Arguably, for each clean image $x$, we could select a $x_{\text{guide}}$ for each of the incorrect classes. In other words, untargeted attacks can be achieved by iterating all possible wrong classes using a targeted attack. However, as our primary focus is targeted attacks, we did not test this strategy. It is also worth noting that this could take much longer training and generation time depending on the number of classes.

\subsection{Training Objectives}

By incorporating an additional guiding image $x_{\text{guide}}$, we can design more effective training objectives by imposing constraints on adversarial example $x_{\text{adv}}$.
For targeted attacks, we establish constraints not only between the adversarial example $x_{\text{adv}}$ and the clean image $x$ but also between the adversarial example $x_{\text{adv}}$ and the guiding image $x_{\text{guide}}$.
These constraints allow us to generate adversarial examples that can effectively mislead the model into predicting a specified target label. For untargeted attacks, introducing the guiding image $x_{\text{guide}}$ can help reduce the risk of overfitting to the clean image $x$, thus making the adversarial example more transferable across different models and data domains.

\subsubsection{Targeted Attack}
For targeted attacks, the objective is to generate adversarial examples that can mislead the model into predicting a target label. Intuitively, in order to fool the surrogate model into predicting the desired target label, the predicted logit values of the adversarial example $x_{\text{adv}}$ must be close to the target label.
Meanwhile, the adversarial example should be able to prevent the surrogate model from perceiving the original content in the clean image $x$. To achieve these two goals, we propose the following contrastive logits loss:
\begin{align}\label{eq:tlc}
	\mathcal{L}_{\text{tlc}} & = \frac{1}{2} { \left[ f \left( x_{\text{adv}}, \theta_{c} \right) - f \left( x_{\text{guide}}, \theta_{c} \right) \right] }^{2} \notag                          \\
	                         & +  \frac{1}{2} { \left[ \max \left( 0, m - \left\| f \left( x_{\text{adv}}, \theta_{c} \right) - f \left( x, \theta_{c} \right) \right\|  \right) \right] }^{2},
\end{align}
where $f \left( \cdot, \theta_{c} \right)$ represents the logit output of the surrogate model, $x_{\text{guide}}$ denotes the guiding image that is of the target class, $m$ is a margin hyperparameter controlling the degree of separation between the two logit vectors. In our experiments, we set the default value of $m$ to 0.2.

The above logit loss may still have the overfitting issue as the logits are closely related to the decision boundary of the surrogate model. It is thus crucial to ensure that the mid-layer features of the adversarial example are also close to those of the target class images.
In the meantime, these mid-layer features must remain substantially distant from those of the source image $x$. To achieve these two objectives, we propose the following enhanced similarity loss:
\begin{align} \label{eq:tfs}
	\mathcal{L}_{\text{tfs}} & = \mathcal{L}_{\text{cos}} \left( f^{l}(x_{\text{adv}}, \theta_{c}), f^{l}(x, \theta_{c}) \right) \notag                                      \\
	                         & - \mathcal{L}_{\text{cos}} \left( f^{l} \left( x_{\text{adv}}, \theta_{c} \right), f^{l} \left( x_{\text{guide}}, \theta_{c} \right) \right),
\end{align}
where $\mathcal{L}_{\text{cos}}$ is the cosine similarity loss.

Combining the above two losses yields the total loss used to train the generator $G((x, x_{\text{guide}}), \theta_{g})$:
\begin{equation}
	\label{eq:tall}
	\small
	\mathcal{L}_{\text{tar}} = \mathcal{L}_{\text{tlc}} + \mathcal{L}_{\text{tfs}}.
\end{equation}
Note that the adversarial example that appears in each of the two loss terms is the output of the generator: $x_{\text{adv}} = G((x, x_{\text{guide}}), \theta_{g})$.

\subsubsection{Untargeted Attack}
The traditional design of untargeted attacks does not have a target label and its sole purpose is to incur errors in the surrogate model. Following previous works, this purpose can be achieved by feature disruption. Therefore, we employ the cosine similarity loss to enforce the adversarial features to be far away from the clean features:
\begin{equation}
	\label{eq:ufs}
	\small
	\mathcal{L}_{\text{ufs}} = \mathcal{L}_{\text{cos}} \left( f^{l} \left( x_{\text{adv}}, \theta_{c} \right), f^{l} \left( x, \theta_{c} \right) \right),
\end{equation}
where $\mathcal{L}_{\text{cos}}$ is the cosine similarity loss. This loss ensures that the adversarial perturbation should be able to cause errors in the feature space. As the features are distorted not necessarily towards a certain target class, this is a suitable base loss for untargeted attacks.

To improve transferability, we define the following untargeted semantic injection loss to incorporate the semantics information from $N$ guiding images:
\begin{equation}
	\label{eq:usi}
	\small
	\mathcal{L}_{\text{usi}} = \frac{1}{N} \sum_{1}^{N} \mathcal{L}_\text{cos} \left( f^{l} \left( x_{\text{guide}}^{i}, \theta_{c} \right), f^{l} \left( x, \theta_{c} \right) \right),
\end{equation}
where $N$ denotes the number of random selections which is set to $N = 16$ in our experiments. Note that all the $N$ selected guiding images are of the same class that is different from the correct class of $x$. The above loss improves transferability by forcing the generated adversarial examples close to the semantics of a randomly chosen untarget (and incorrect) class.
As we explained earlier, we did not explore the more time-consuming version of our attack that iterates over all possible incorrect classes.

The overall training objective for untargeted attacks can be formulated as:
\begin{equation}
	\label{eq:uall}
	\small
	\mathcal{L}_{\text{untar}} =\mathcal{L}_{\text{ufs}} + \mathcal{L}_{\text{usi}}.
\end{equation}

After training, the generator $G \left( \left( \cdot, \cdot \right), \theta_{g} \right)$ can be used to generate an adversarial example for any given clean image via a single forward pass. It is worth noting that the generation process also requires the guiding image $x_{guide}$. Due to the generalizability of the generator model, the adversary can select the guiding image following the same procedure as used during training and obtain completely different guiding images for the test images. We also note that the guiding images used for generation are also different from those used for training the generator.

\section{Experiments}
\label{sec:exp}

\subsection{Experimental Setup}

\paragraph{Datasets and Models}
We evaluate our method for three different settings, as follows:

\begin{itemize}
	\item \textbf{Targeted and untargeted cross-architecture attacks:}~We train the adversarial generator using the ImageNet dataset~\cite{imagenet} with the feedbacks from surrogate models.
	      For targeted attack scenarios, we employ architectures noted for their high transferability, specifically ResNet152~\cite{resnet} and DenseNet169~\cite{densenet}.
	      Conversely, we select less robust architectures, such as VGG-16~\cite{vgg}, for untargeted attack assessments.
	      In the evaluation phase, we analyze the transferability of the generated adversarial examples across ten distinct model architectures: VGG-16, VGG-19, ResNet-50, ResNet-152, DenseNet-121, DenseNet-169, Inception-V3, ViT-B/16~\cite{vit}, ViT-B/32, and Swin-B~\cite{swin}. The model weights are obtained from the \texttt{Torchvision} model zoo.

	\item \textbf{Untargeted cross-domain attacks:}~Similar to cross-architecture settings, we train the adversarial generator using the ImageNet dataset, with VGG-16 as the surrogate model.
	      Unlike previous methods that focused solely on cross-architecture settings, we follow the literature~\citet{bia-attack} to assess adversarial transferability across three distinct datasets: CUB-200~\cite{cub}, Stanford Cars~\cite{cars}, and Oxford Flowers~\cite{flowers}.
	      For these evaluations, we employ three different model architectures: ResNet-50, SENet-154~\cite{senet}, and SE-ResNet-101, all pre-trained using the DCL framework~\cite{dcl}.
\end{itemize}

\paragraph{Evaluation Metrics and Baselines}

We evaluate our method using the top-1 classification accuracy.
For targeted attacks, we train adversarial generators for three classes: \texttt{Great Grey Owl} (class No. 24), \texttt{Goose} (class No. 99), and \texttt{French Bulldog} (class No. 245), and report the average top-1 accuracy across these classes, following the literature~\cite{ttaa-attack}.
In contrast, for untargeted attacks, we provide the average top-1 accuracy across all classes.

For better comparison, we select the state-of-the-art methods from iterative and generative attacks as our baselines.
Firstly, we select two logits based methods(PGD~\cite{pgd-attack}, DI-FGSM~\cite{dim-attack}) and one mid-layer feature based method(DR~\cite{dr-attack}) as our iterative baseline methods.
Secondly, for generative attacks, we select CDA~\cite{cda-attack} and BIA~\cite{bia-attack} as our baseline competitors.
Additionally, for targeted settings, we also incorporate GAP~\cite{gap-attack} and TTP~\cite{ttp-attack} as competitors within generative methodologies.
It is worth noting that we omit BIA~\cite{bia-attack} in untargeted settings because of its limited objective design.

\subsubsection{Implementation Details}~In this framework, we utilize the ResNet generator as the base adversarial generator.
The training process employs the Adam optimizer with a learning rate of $2e^{-4}$. We set momentum decay factors at 0.5 and 0.999.
We train the generator for 1 epoch with a batch size of 16.
To extract layer features from the surrogate model, we adopt the mid-layer selection used in BIA~\cite{bia-attack}.
As for the attack settings, we establish the following parameters: an attack budget of $\epsilon=16/255$ for targeted settings and $\epsilon=10/255$ for untargeted settings. For iterative attacks, we follow the same configurations as \citet{bia-attack}.

\subsection{Experimental Results}

\begin{table*}[!t]
	\small
	\centering
	\begin{tabular}{ccccccccccccccc}
		\toprule
		\multirow{1}{*}{Sur.} & Attack
		                      & V16      & V19            & R50            & R152
		                      & D121     & D169           & Inc
		                      & VB/16    & VB/32          & Swin/B
		                      & Avg/Conv & Avg/ViT        & Avg/All                                                                                                                                                                                                  \\
		\midrule \multirow{6}{*}{\rotatebox{90}{R152}}
		                      & PGD      & 0.78           & 0.63           & 7.96           & 93.56          & 2.11           & 2.15           & 0.44           & 0.08           & 0.03          & 0.12           & 15.38          & 0.08           & 10.79          \\
		                      & DI-FGSM  & 4.85           & 4.20           & 34.84          & 95.34          & 23.71          & 25.45          & 6.17           & 0.48           & 0.12          & 0.96           & 27.79          & 0.52           & 19.61          \\
		\cmidrule{2-15}
		                      & CDA      & 19.46          & 19.92          & 71.68          & 95.74          & 78.58          & 71.12          & 27.05          & 3.39           & 0.62          & 5.83           & 54.79          & 3.28           & 39.34          \\
		                      & GAP      & 29.38          & 27.28          & 76.46          & 95.11          & 72.21          & 70.68          & 13.98          & 4.53           & 0.76          & 6.21           & 55.01          & 3.83           & 39.66          \\
		                      & TTP      & 29.87          & 22.97          & 82.29          & 97.60          & 80.13          & 71.20          & 24.56          & 5.50           & 0.28          & 11.71          & 58.38          & 5.83           & 42.61          \\
		                      & Ours     & \textbf{75.11} & \textbf{73.38} & \textbf{87.35} & \textbf{97.86} & \textbf{82.15} & \textbf{81.79} & \textbf{54.63} & \textbf{25.23} & \textbf{8.24} & \textbf{29.38} & \textbf{78.90} & \textbf{20.95} & \textbf{61.51} \\
		\midrule \multirow{6}{*}{\rotatebox{90}{D169}}
		                      & PGD      & 1.37           & 1.12           & 5.03           & 2.13           & 10.71          & 97.94          & 0.53           & 0.06           & 0.02          & 0.21           & 16.97          & 0.10           & 11.91          \\
		                      & DI-FGSM  & 5.35           & 4.79           & 20.31          & 11.05          & 43.00          & 98.25          & 5.64           & 0.43           & 0.09          & 0.83           & 26.91          & 0.45           & 18.97          \\
		\cmidrule{2-15}
		                      & CDA      & 11.48          & 15.34          & 43.01          & 35.82          & 63.41          & 95.46          & 18.23          & 2.25           & 0.30          & 3.15           & 40.39          & 1.90           & 28.85          \\
		                      & GAP      & 4.54           & 8.20           & 15.91          & 13.18          & 49.96          & 64.79          & 8.70           & 1.11           & 0.39          & 1.69           & 23.61          & 1.06           & 16.85          \\
		                      & TTP      & 39.00          & 33.18          & 64.71          & 46.74          & 90.23          & 97.62          & 17.34          & 8.76           & 0.55          & 8.39           & 55.55          & 5.90           & 40.65          \\
		                      & Ours     & \textbf{76.32} & \textbf{77.14} & \textbf{78.44} & \textbf{67.69} & \textbf{93.11} & \textbf{97.84} & \textbf{48.53} & \textbf{30.61} & \textbf{8.61} & \textbf{33.96} & \textbf{77.01} & \textbf{24.39} & \textbf{61.23} \\
		\bottomrule
	\end{tabular}
	\caption{Evaluation results for targeted cross-model attack. We report the top-1 average accuracy for the three targeted labels, with higher accuracy indicative of improved performance. The perturbation budget is constrained by $\left\|x_{\text{adv}} - x\right\|_{\infty} \leq 16/255$.}
	\label{tab:tar-cra}
\end{table*}

\paragraph{Targeted Cross-architecture Transferability}
In Table~\ref{tab:tar-cra}, we present the average top-1 accuracy results across three targeted classes, wherein higher values indicate superior performance. The best-performing results are delineated using bold formatting. The first column (Sur.) shows the surrogate models and the first row corresponds to different target models.
It is evident that our methodology consistently surpasses all alternative approaches across every model, achieving significant improvements. For instance, utilizing DenseNet-169 as the surrogate model, we attain an average accuracy enhancement of 32.38\% in comparison to the baseline method, CDA. This trend is similarly observed across all models, further emphasizing the remarkable effectiveness of our approach in augmenting adversarial transferability.
What's more, take a look at the transferability comparison with other methods on ViT target models, previous methods achieved only negligible attack effectiveness. For example, the most effective method, TTP, which employs DenseNet169 as the surrogate model, attained an average attack success rate of just 5.90\%. In contrast, our method is the first to successfully conduct attacks on the ViT architecture, achieving a significantly higher attack success rate of 24.39\%.

\begin{table*}[!t]
	\small
	\centering
	\begin{tabular}{cccccccccccc}
		\toprule
		\multirow{3}{*}{Sur.} & \multirow{2}{*}{Attacks} & \multicolumn{3}{c}{CUB-200-2011} & \multicolumn{3}{c}{Stanford Cars} & \multicolumn{3}{c}{FGVC Aircraft} & \multirow{2}{*}{Avg/All}                                                                                                       \\
		                      &                          & R50                              & SE154                             & SE-R101                           & R50                      & SE154          & SE-R101        & R50            & SE154          & SE-R101                         \\
		\cmidrule{2-12}
		                      & Clean                    & 87.23                            & 86.30                             & 85.88                             & 90.34                    & 89.45          & 89.17          & 71.08          & 71.56          & 73.84          & 82.76          \\
		\toprule \multirow{6}{*}{\rotatebox{90}{V16}}
		                      & PGD                      & 81.62                            & 80.19                             & 80.91                             & 78.81                    & 82.43          & 84.90          & 51.16          & 55.30          & 58.00          & 72.59          \\
		                      & DI-FGSM                  & 80.19                            & 76.94                             & 78.65                             & 76.45                    & 77.54          & 82.76          & 48.27          & 51.76          & 53.74          & 69.59          \\
		                      & DR                       & 80.15                            & 81.53                             & 81.72                             & 82.69                    & 86.26          & 86.56          & 54.28          & 62.62          & 61.99          & 75.31          \\
		\cmidrule{2-12}
		                      & CDA                      & 67.69                            & 60.58                             & 70.43                             & 50.90                    & 46.05          & 69.10          & 35.88          & 31.59          & 39.87          & 52.45          \\
		                      & BIA                      & 65.24                            & 63.05                             & 67.78                             & 53.07                    & 51.04          & 62.93          & 34.89          & 36.60          & 39.63          & 52.69          \\
		                      & ours                     & \textbf{48.41}                   & \textbf{40.56}                    & \textbf{56.08}                    & \textbf{42.37}           & \textbf{32.31} & \textbf{54.51} & \textbf{30.18} & \textbf{25.41} & \textbf{33.21} & \textbf{40.34} \\
		\bottomrule
	\end{tabular}
	\caption{Evaluation results for untargeted cross-domain attacks. The perturbation generators have been trained to utilize the ImageNet data domain in conjunction with the surrogate model VGG-16. We report the top-1 average accuracy, wherein a lower value signifies better performance. The perturbation budget is constrained by $\left\|x_{\text{adv}} - x\right\|_{\infty} \leq 10/255$.}
	\label{tab:untar-crd}
\end{table*}

\paragraph{Untargeted Transferability}
We further investigate the untargeted transferability within the cross-domain and the cross-architecture scenario.
Firstly, in Table~\ref{tab:untar-crd}, we present the average top-1 accuracy results for the untargeted cross-domain attacks. The best results are highlighted using bold formatting. Our method consistently outperforms all competing approaches. Notably, our approach achieves an average accuracy of 40.34\%, significantly surpassing BIA.
Secondly, in Table~\ref{tab:untar-cra}, we present the average top-1 accuracy results for the untargeted cross-architecture attacks. Our method achieves a lower average top-1 accuracy of 38.78\% (lower for better performance). A closer inspection reveals that our approach demonstrates relatively better performance across different architectures, outperforming other methods in both convolutional and transformer models.

\begin{table*}[!t]
	\small
	\centering
	\begin{tabular}{ccccccccccccccccc}
		\toprule
		\multirow{2}{*}{Sur.} & Attack
		                      & V16      & V19           & R50           & R52
		                      & D121     & D169          & Inc
		                      & VB/16    & VB/32         & Swin/B
		                      & Avg/Conv & Avg/ViT       & Avg/All                                                                                                                                                                                                  \\
		\cmidrule{2-15}
		                      & Clean    & 71.58         & 72.40         & 76.15          & 78.33          & 74.43          & 75.58          & 69.53          & 81.07          & 75.91          & 83.17          & 74.00          & 80.05          & 75.82          \\
		\midrule \multirow{6}{*}{\rotatebox{90}{V16}}
		                      & PGD      & \textbf{0.07} & \textbf{0.82} & 35.30          & 48.80          & 36.70          & 41.91          & 52.51          & 73.09          & 71.19          & \textbf{64.50} & 30.87          & \textbf{69.59} & 42.49          \\
		                      & DI-FGSM  & 0.07          & 1.14          & 39.59          & 52.55          & 41.31          & 46.42          & 54.74          & 74.06          & 71.86          & 66.83          & 33.69          & 70.92          & 44.86          \\
		                      & DR       & 14.71         & 35.98         & 62.20          & 68.20          & 61.68          & 65.28          & 61.24          & 75.34          & 70.98          & 75.34          & 52.76          & 73.89          & 59.10          \\
		\cmidrule{2-15}
		                      & CDA      & 12.58         & 19.82         & 49.40          & 58.52          & 49.14          & 54.55          & 50.60          & 74.36          & 70.92          & 73.36          & 42.09          & 72.88          & 51.33          \\
		                      & BIA      & 1.16          & 2.59          & 44.96          & 53.82          & 44.10          & 49.60          & 51.97          & 73.97          & \textbf{69.96} & 70.23          & 35.46          & 71.39          & 46.24          \\
		                      & ours     & 1.17          & 3.28          & \textbf{26.40} & \textbf{43.73} & \textbf{27.31} & \textbf{33.01} & \textbf{42.36} & \textbf{72.40} & 70.65          & 67.52          & \textbf{25.32} & 70.19          & \textbf{38.78} \\
		\bottomrule
	\end{tabular}
	\caption{Evaluation results for untargeted cross-architecture attacks. The perturbation generators have been trained to utilize the ImageNet data domain in conjunction with surrogate models, specifically VGG-16. We report the top-1 average accuracy, wherein a lower value signifies better performance. The perturbation budget is constrained by $\left\|x_{\text{adv}} - x\right\|_{\infty} \leq 10/255$.}
	\label{tab:untar-cra}
\end{table*}

\begin{figure}[!t]
	\centering
	\begin{tabular}{@{}c@{ }c@{ }c@{ }c@{ }c@{}}
		$x$
		 & $x_{\text{guide}}$
		 & $x_{\text{adv}}$/CDA
		 & $x_{\text{adv}}$/TTP
		 & $x_{\text{adv}}$/Ours                                                                                      \\
		\includegraphics[width=0.09\textwidth]{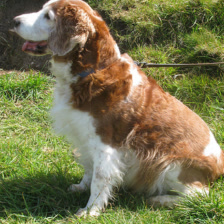}
		 & \includegraphics[width=0.09\textwidth]{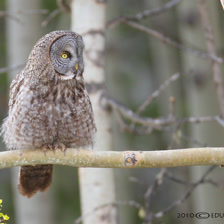}
		 & \includegraphics[width=0.09\textwidth]{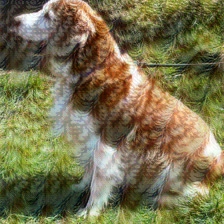}
		 & \includegraphics[width=0.09\textwidth]{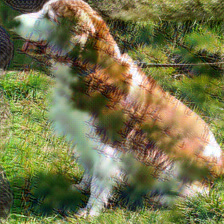}
		 & \includegraphics[width=0.09\textwidth]{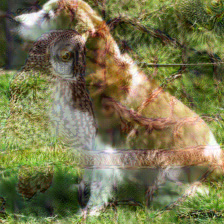}    \\
		\includegraphics[width=0.09\textwidth]{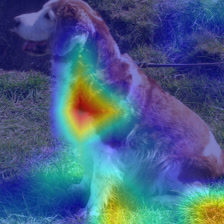}
		 & \includegraphics[width=0.09\textwidth]{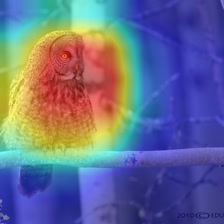}
		 & \includegraphics[width=0.09\textwidth]{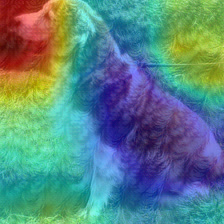}
		 & \includegraphics[width=0.09\textwidth]{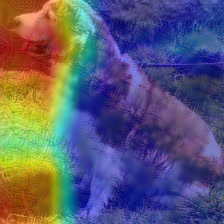}
		 & \includegraphics[width=0.09\textwidth]{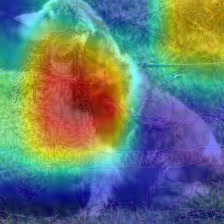}
	\end{tabular}
	\caption{Illustration of attention shift. We use Grad-CAM visualization of adversarial examples in the targeted attack setting. The adversarial examples were generated using ResNet-152 as the surrogate model, with evaluations conducted on ResNet-50 as the target model.}
	\label{fig:vis-gc}
\end{figure}

\paragraph{Visualization of Targeted Attacks}
In Figure~\ref{fig:vis-gc}, we illustrate visualizations of adversarial examples generated by our methodology in the context of targeted attacks. Using Grad-CAM~\cite{gradcam}, we highlight the regions of interest in the natural input image $x$, the guidance image $x_{\text{guide}}$, and the generated adversarial examples $x_{\text{adv}}$ for two generative attack methods (CDA, TTP) and our method.
Firstly, take a look at the first row, the results demonstrate the effectiveness of our approach in creating adversarial examples that are visually indistinguishable from their corresponding natural images.
Secondly, the results in the second row illustrate the effectiveness of the additional guidance image in shifting attention. Our methods generate adversarial examples that align more closely with the semantics of the guidance image. This shows that incorporating additional image guidance allows for a more controlled attack, both positionally and semantically.

\begin{table}[!t]
	\small
	\centering
	\begin{tabular}{cc}
		\begin{tabular}{ccccc}
			\toprule
			\multicolumn{4}{c}{Targeted Attack}                                             \\
			\midrule
			Sur. & $\mathcal{L}_{\text{tlc}}$ & $\mathcal{L}_{\text{tfs}}$ & Arch           \\
			\midrule
			\multirow{3}{*}{\rotatebox{90}{R152}}
			     & \checkmark                 & -                          & 16.51          \\
			     & -                          & \checkmark                 & 42.08          \\
			     & \checkmark                 & \checkmark                 & \textbf{61.51} \\
			\bottomrule
		\end{tabular} &
		\begin{tabular}{ccccc}
			\toprule
			\multicolumn{5}{c}{Untargeted Attack}                                                            \\
			\midrule
			Sur. & $\mathcal{L}_{\text{ufs}}$ & $\mathcal{L}_{\text{usi}}$ & Dom            & Arch           \\
			\midrule
			\multirow{3}{*}{\rotatebox{90}{V16}}
			     & \checkmark                 & -                          & 50.48          & 44.67          \\
			     & -                          & \checkmark                 & 65.62          & 47.70          \\
			     & \checkmark                 & \checkmark                 & \textbf{40.34} & \textbf{38.78} \\
			\bottomrule
		\end{tabular} \\
	\end{tabular}
	\caption{Ablation study on loss objectives. The left table presents the average top-1 accuracy associated with various objective functions in an targeted scenario (notably, lower values are preferable), whereas the right table details the results obtained under untargeted configurations (wherein higher values are desirable).}
	\label{tab:abla-loss}
\end{table}

\paragraph{Objective Functions Ablation}
Table~\ref{tab:abla-loss} presents the findings from the ablation study concerning the loss objectives. In the domain of untargeted attacks, our observations indicate that the integration of the semantic injection loss $\mathcal{L}_{\text{ufs}}$ relatively enhances performance. However, it is the semantic injection loss $\mathcal{L}_{\text{usi}}$ that plays a more pivotal role in mitigating overfitting during training, thereby enhancing adversarial transferability. Conversely, in the context of targeted attacks, both the logits contrastive loss $\mathcal{L}_{\text{lc}}$ and the similarity loss $\mathcal{L}_{\text{tfs}}$ emerge as critical components. The results demonstrate that the synergistic application of these two losses can substantially elevate adversarial transferability.

\paragraph{Guiding Image Selection Strategy}
Table~\ref{tab:guid-selection} presents the results from the ablation study concerning the guiding image selection strategies. We formulate two more strategies: 1) a CLIP-score based selection, which selects the image with the maximum or minimum CLIP score, and 2) manual selection, which selects images that are highly representative of the target class. Using clip-score is less effective than random selection, primarily due to insufficient consideration of the overlap between the target category and the guiding image. In contrast, high-quality manual selection can achieve better performance than random selection.

\begin{table}[!t]
	\small
	\centering
	\begin{tabular}{ccccc}
		\toprule
		Strategy               & Avg/Conv & Avg/ViT & Avg/All \\
		\midrule
		Random                 & 74.56    & 16.55   & 57.16   \\
		CLIP-Score~(min)       & 66.77    & 13.57   & 50.81   \\
		CLIP-Score~(max)       & 67.27    & 13.71   & 51.20   \\
		Manual~(\texttt{15})   & 80.45    & 20.78   & 62.55   \\
		Manual~(\texttt{3861}) & 82.49    & 24.58   & 65.11   \\
		\bottomrule
	\end{tabular}
	\caption{Ablation study on guiding image selection strategies. The results are presented in terms of average top-1 accuracy, with higher values indicating superior performance. The guiding image selection strategies are evaluated on \texttt{Great Grey Owl}~(No.~24).}
	\label{tab:guid-selection}
\end{table}

\paragraph{Computational Analysis of the Semantic Injection Module}
In Table~\ref{tab:sim-comp}, we present a computational analysis of the semantic injection module. The results are presented in terms of the number of parameters, FLOPs, and average time required for generating one adversarial examples. The results indicate that the semantic injection module incurs a slight increase in computational overhead, with the average time required for generating adversarial examples increasing from 3.0 ms to 7.8 ms. However, the additional computational cost is justified by the substantial improvements in adversarial transferability.

\begin{table}[!t]
	\small
	\centering
	\begin{tabular}{ccccc}
		\toprule
		          & Params~(M)   & FLOPs(G)      & Avg Time~(ms) \\
		\midrule
		PGD       & -            & -             & 543.5         \\
		CDA       & $3.25~e^{4}$ & $0.78~e^{-2}$ & 3.0           \\
		CDA + SIM & $7.92~e^{4}$ & $1.64~e^{-2}$ & 7.8           \\
		\bottomrule
	\end{tabular}
	\caption{Computational analysis of the semantic injection module. The results are presented in terms of the number of parameters, FLOPs, and average time required for generating adversarial examples.}
	\label{tab:sim-comp}
\end{table}

\section{Conclusion}
\label{sec:conclu}

We introduce a new framework that uses additional image guidance for targeted and untargeted transferable attacks. A semantic injection module is integrated into a base adversarial generator to improve the generation of transferable adversarial examples. We also propose innovative loss objectives to enhance the guidance for adversarial generation. Extensive experiments show our method significantly improves adversarial transferability, outperforming state-of-the-art techniques.

\section*{Acknowledgments}

This work is in part supported by the National Key R\&D Program of China (Grant No. 2021ZD0112804) and the National Natural Science Foundation of China (Grant No. 62276067).

\bibliography{aim}

\begin{thebibliography}{40}
\providecommand{\natexlab}[1]{#1}

\bibitem[{Aich et~al.(2022)Aich, Ta, Gupta, Song, Krishnamurthy, Asif, and
  Roy-Chowdhury}]{gama-attack}
Aich, A.; Ta, C.-K.; Gupta, A.; Song, C.; Krishnamurthy, S.; Asif, S.; and
  Roy-Chowdhury, A. 2022.
\newblock Gama: Generative adversarial multi-object scene attacks.
\newblock \emph{NeurIPS}, 35: 36914--36930.

\bibitem[{Baluja and Fischer(2017)}]{atn-attack}
Baluja, S.; and Fischer, I. 2017.
\newblock Adversarial transformation networks: Learning to generate adversarial
  examples.
\newblock \emph{arXiv preprint arXiv:1703.09387}.

\bibitem[{Chen et~al.(2022)Chen, Wei, Chen, Wu, and Jiang}]{bsc-attack}
Chen, K.; Wei, Z.; Chen, J.; Wu, Z.; and Jiang, Y.-G. 2022.
\newblock Attacking video recognition models with bullet-screen comments.
\newblock In \emph{AAAI}, volume~36, 312--320.

\bibitem[{Chen et~al.(2023)Chen, Wei, Chen, Wu, and Jiang}]{gcma-attack}
Chen, K.; Wei, Z.; Chen, J.; Wu, Z.; and Jiang, Y.-G. 2023.
\newblock GCMA: Generative Cross-Modal Transferable Adversarial Attacks from
  Images to Videos.
\newblock In \emph{ACM MM}, 698--708.

\bibitem[{Chen et~al.(2019)Chen, Bai, Zhang, and Mei}]{dcl}
Chen, Y.; Bai, Y.; Zhang, W.; and Mei, T. 2019.
\newblock Destruction and construction learning for fine-grained image
  recognition.
\newblock In \emph{CVPR}, 5157--5166.

\bibitem[{Deng et~al.(2009)Deng, Dong, Socher, Li, Li, and Fei-Fei}]{imagenet}
Deng, J.; Dong, W.; Socher, R.; Li, L.-J.; Li, K.; and Fei-Fei, L. 2009.
\newblock Imagenet: A large-scale hierarchical image database.
\newblock In \emph{CVPR}, 248--255.

\bibitem[{Dosovitskiy et~al.(2020)Dosovitskiy, Beyer, Kolesnikov, Weissenborn,
  Zhai, Unterthiner, Dehghani, Minderer, Heigold, Gelly et~al.}]{vit}
Dosovitskiy, A.; Beyer, L.; Kolesnikov, A.; Weissenborn, D.; Zhai, X.;
  Unterthiner, T.; Dehghani, M.; Minderer, M.; Heigold, G.; Gelly, S.; et~al.
  2020.
\newblock An image is worth 16x16 words: Transformers for image recognition at
  scale.
\newblock \emph{arXiv preprint arXiv:2010.11929}.

\bibitem[{Goodfellow, Shlens, and Szegedy(2015)}]{fgsm-attack}
Goodfellow, I.~J.; Shlens, J.; and Szegedy, C. 2015.
\newblock Explaining and Harnessing Adversarial Examples.
\newblock \emph{arXiv preprint arXiv:1412.6572}.

\bibitem[{He et~al.(2016)He, Zhang, Ren, and Sun}]{resnet}
He, K.; Zhang, X.; Ren, S.; and Sun, J. 2016.
\newblock Deep residual learning for image recognition.
\newblock In \emph{CVPR}, 770--778.

\bibitem[{Hochreiter and Schmidhuber(1997)}]{lstm}
Hochreiter, S.; and Schmidhuber, J. 1997.
\newblock Long short-term memory.
\newblock \emph{Neural computation}, 9(8): 1735--1780.

\bibitem[{Hu, Shen, and Sun(2018)}]{senet}
Hu, J.; Shen, L.; and Sun, G. 2018.
\newblock Squeeze-and-excitation networks.
\newblock In \emph{CVPR}, 7132--7141.

\bibitem[{Huang et~al.(2017)Huang, Liu, Van Der~Maaten, and
  Weinberger}]{densenet}
Huang, G.; Liu, Z.; Van Der~Maaten, L.; and Weinberger, K.~Q. 2017.
\newblock Densely connected convolutional networks.
\newblock In \emph{CVPR}, 4700--4708.

\bibitem[{Krause et~al.(2013)Krause, Stark, Deng, and Fei-Fei}]{cars}
Krause, J.; Stark, M.; Deng, J.; and Fei-Fei, L. 2013.
\newblock 3d object representations for fine-grained categorization.
\newblock In \emph{ICCV workshops}, 554--561.

\bibitem[{Krizhevsky, Sutskever, and Hinton(2012)}]{alexnet}
Krizhevsky, A.; Sutskever, I.; and Hinton, G.~E. 2012.
\newblock Imagenet classification with deep convolutional neural networks.
\newblock \emph{NeurIPS}, 25.

\bibitem[{Kurakin, Goodfellow, and Bengio(2018)}]{ifgsm-attack}
Kurakin, A.; Goodfellow, I.~J.; and Bengio, S. 2018.
\newblock Adversarial examples in the physical world.
\newblock In \emph{Artificial intelligence safety and security}, 99--112.

\bibitem[{LeCun et~al.(1998)LeCun, Bottou, Bengio, and Haffner}]{lenet}
LeCun, Y.; Bottou, L.; Bengio, Y.; and Haffner, P. 1998.
\newblock Gradient-based learning applied to document recognition.
\newblock \emph{Proceedings of the IEEE}, 86(11): 2278--2324.

\bibitem[{Li et~al.(2020)Li, Deng, Li, Yan, Gao, and Huang}]{ttta-attack}
Li, M.; Deng, C.; Li, T.; Yan, J.; Gao, X.; and Huang, H. 2020.
\newblock Towards transferable targeted attack.
\newblock In \emph{CVPR}, 641--649.

\bibitem[{Li et~al.(2024)Li, Wang, Li, Chen, and Zhang}]{ucg-attack}
Li, Z.; Wang, W.; Li, J.; Chen, K.; and Zhang, S. 2024.
\newblock UCG: A Universal Cross-Domain Generator for Transferable Adversarial
  Examples.
\newblock \emph{IEEE TIFS}.

\bibitem[{Li et~al.(2023)Li, Wu, Su, Zheng, and Lyu}]{cdta-attack}
Li, Z.; Wu, W.; Su, Y.; Zheng, Z.; and Lyu, M.~R. 2023.
\newblock CDTA: a cross-domain transfer-based attack with contrastive learning.
\newblock In \emph{AAAI}, volume~37 of \emph{2}, 1530--1538.

\bibitem[{Liu et~al.(2021)Liu, Lin, Cao, Hu, Wei, Zhang, Lin, and Guo}]{swin}
Liu, Z.; Lin, Y.; Cao, Y.; Hu, H.; Wei, Y.; Zhang, Z.; Lin, S.; and Guo, B.
  2021.
\newblock Swin transformer: Hierarchical vision transformer using shifted
  windows.
\newblock In \emph{ICCV}, 10012--10022.

\bibitem[{Lu et~al.(2020)Lu, Jia, Wang, Li, Chai, Carin, and
  Velipasalar}]{dr-attack}
Lu, Y.; Jia, Y.; Wang, J.; Li, B.; Chai, W.; Carin, L.; and Velipasalar, S.
  2020.
\newblock Enhancing cross-task black-box transferability of adversarial
  examples with dispersion reduction.
\newblock In \emph{CVPR}, 940--949.

\bibitem[{Madry et~al.(2017)Madry, Makelov, Schmidt, Tsipras, and
  Vladu}]{pgd-attack}
Madry, A.; Makelov, A.; Schmidt, L.; Tsipras, D.; and Vladu, A. 2017.
\newblock Towards deep learning models resistant to adversarial attacks.
\newblock \emph{arXiv preprint arXiv:1706.06083}.

\bibitem[{Naseer et~al.(2021)Naseer, Khan, Hayat, Khan, and
  Porikli}]{ttp-attack}
Naseer, M.; Khan, S.; Hayat, M.; Khan, F.~S.; and Porikli, F. 2021.
\newblock On generating transferable targeted perturbations.
\newblock In \emph{ICCV}, 7708--7717.

\bibitem[{Naseer et~al.(2019)Naseer, Khan, Khan, Shahbaz~Khan, and
  Porikli}]{cda-attack}
Naseer, M.~M.; Khan, S.~H.; Khan, M.~H.; Shahbaz~Khan, F.; and Porikli, F.
  2019.
\newblock Cross-domain transferability of adversarial perturbations.
\newblock \emph{NeurIPS}, 32.

\bibitem[{Nilsback and Zisserman(2008)}]{flowers}
Nilsback, M.-E.; and Zisserman, A. 2008.
\newblock Automated flower classification over a large number of classes.
\newblock In \emph{Indian conference on computer vision, graphics \& image
  processing}, 722--729.

\bibitem[{Park et~al.(2019)Park, Liu, Wang, and Zhu}]{spade}
Park, T.; Liu, M.-Y.; Wang, T.-C.; and Zhu, J.-Y. 2019.
\newblock Semantic image synthesis with spatially-adaptive normalization.
\newblock In \emph{CVPR}, 2337--2346.

\bibitem[{Poursaeed et~al.(2018)Poursaeed, Katsman, Gao, and
  Belongie}]{gap-attack}
Poursaeed, O.; Katsman, I.; Gao, B.; and Belongie, S. 2018.
\newblock Generative adversarial perturbations.
\newblock In \emph{CVPR}, 4422--4431.

\bibitem[{Radford et~al.(2021)Radford, Kim, Hallacy, Ramesh, Goh, Agarwal,
  Sastry, Askell, Mishkin, Clark et~al.}]{clip}
Radford, A.; Kim, J.~W.; Hallacy, C.; Ramesh, A.; Goh, G.; Agarwal, S.; Sastry,
  G.; Askell, A.; Mishkin, P.; Clark, J.; et~al. 2021.
\newblock Learning transferable visual models from natural language
  supervision.
\newblock In \emph{ICML}, 8748--8763.

\bibitem[{Redmon et~al.(2016)Redmon, Divvala, Girshick, and Farhadi}]{yolo}
Redmon, J.; Divvala, S.; Girshick, R.; and Farhadi, A. 2016.
\newblock You only look once: Unified, real-time object detection.
\newblock In \emph{CVPR}, 779--788.

\bibitem[{Rombach et~al.(2022)Rombach, Blattmann, Lorenz, Esser, and
  Ommer}]{stable-diffusion}
Rombach, R.; Blattmann, A.; Lorenz, D.; Esser, P.; and Ommer, B. 2022.
\newblock High-resolution image synthesis with latent diffusion models.
\newblock In \emph{CVPR}, 10684--10695.

\bibitem[{Selvaraju et~al.(2017)Selvaraju, Cogswell, Das, Vedantam, Parikh, and
  Batra}]{gradcam}
Selvaraju, R.~R.; Cogswell, M.; Das, A.; Vedantam, R.; Parikh, D.; and Batra,
  D. 2017.
\newblock Grad-cam: Visual explanations from deep networks via gradient-based
  localization.
\newblock In \emph{ICCV}, 618--626.

\bibitem[{Simonyan and Zisserman(2014)}]{vgg}
Simonyan, K.; and Zisserman, A. 2014.
\newblock Very deep convolutional networks for large-scale image recognition.
\newblock \emph{arXiv preprint arXiv:1409.1556}.

\bibitem[{Szegedy et~al.(2013)Szegedy, Zaremba, Sutskever, Bruna, Erhan,
  Goodfellow, and Fergus}]{intriguing}
Szegedy, C.; Zaremba, W.; Sutskever, I.; Bruna, J.; Erhan, D.; Goodfellow, I.;
  and Fergus, R. 2013.
\newblock Intriguing properties of neural networks.
\newblock \emph{arXiv preprint arXiv:1312.6199}.

\bibitem[{Vaswani et~al.(2017)Vaswani, Shazeer, Parmar, Uszkoreit, Jones,
  Gomez, Kaiser, and Polosukhin}]{transformer}
Vaswani, A.; Shazeer, N.; Parmar, N.; Uszkoreit, J.; Jones, L.; Gomez, A.~N.;
  Kaiser, {\L}.; and Polosukhin, I. 2017.
\newblock Attention is all you need.
\newblock \emph{NeurIPS}, 30.

\bibitem[{Wah et~al.(2011)Wah, Branson, Welinder, Perona, and Belongie}]{cub}
Wah, C.; Branson, S.; Welinder, P.; Perona, P.; and Belongie, S. 2011.
\newblock The caltech-ucsd birds-200-2011 dataset.
\newblock Technical Report CNS-TR-2011-001, California Institute of Technology.

\bibitem[{Wang et~al.(2023)Wang, Yang, Feng, Sun, Guo, Zhang, and
  Ren}]{ttaa-attack}
Wang, Z.; Yang, H.; Feng, Y.; Sun, P.; Guo, H.; Zhang, Z.; and Ren, K. 2023.
\newblock Towards transferable targeted adversarial examples.
\newblock In \emph{CVPR}, 20534--20543.

\bibitem[{Xie et~al.(2019)Xie, Zhang, Zhou, Bai, Wang, Ren, and
  Yuille}]{dim-attack}
Xie, C.; Zhang, Z.; Zhou, Y.; Bai, S.; Wang, J.; Ren, Z.; and Yuille, A.~L.
  2019.
\newblock Improving transferability of adversarial examples with input
  diversity.
\newblock In \emph{CVPR}, 2730--2739.

\bibitem[{Yang, Jeong, and Yoon(2024)}]{facl-attack}
Yang, H.; Jeong, J.; and Yoon, K.-J. 2024.
\newblock FACL-Attack: Frequency-Aware Contrastive Learning for Transferable
  Adversarial Attacks.
\newblock In \emph{AAAI}, volume~38 of \emph{6}, 6494--6502.

\bibitem[{Zhang et~al.(2022)Zhang, Li, Chen, Song, Gao, He, and
  Xue}]{bia-attack}
Zhang, Q.; Li, X.; Chen, Y.; Song, J.; Gao, L.; He, Y.; and Xue, H. 2022.
\newblock Beyond imagenet attack: Towards crafting adversarial examples for
  black-box domains.
\newblock \emph{arXiv preprint arXiv:2201.11528}.

\bibitem[{Zhu et~al.(2017)Zhu, Park, Isola, and Efros}]{style-gan}
Zhu, J.-Y.; Park, T.; Isola, P.; and Efros, A.~A. 2017.
\newblock Unpaired image-to-image translation using cycle-consistent
  adversarial networks.
\newblock In \emph{ICCV}, 2223--2232.

\end{thebibliography}

\end{document}